# Flexible Priors for Exemplar-based Clustering

**Daniel Tarlow**
Dept. of Computer Science
University of Toronto
Toronto, ON M5S 3H5

**Richard S. Zemel**
Dept. of Computer Science
University of Toronto
Toronto, ON M5S 3H5

**Brendan J. Frey**
Probabilistic and Statistical Inference Group
University of Toronto
Toronto, ON M5S 3G4

## Abstract

Exemplar-based clustering methods have been shown to produce state-of-the-art results on a number of synthetic and real-world clustering problems. They are appealing because they offer computational benefits over latent-mean models and can handle arbitrary pairwise similarity measures between data points. However, when trying to recover underlying structure in clustering problems, tailored similarity measures are often not enough; we also desire control over the distribution of cluster sizes. Priors such as Dirichlet process priors allow the number of clusters to be unspecified while expressing priors over data partitions. To our knowledge, they have not been applied to exemplar-based models. We show how to incorporate priors, including Dirichlet process priors, into the recently introduced affinity propagation algorithm. We develop an efficient max-product belief propagation algorithm for our new model and demonstrate experimentally how the expanded range of clustering priors allows us to better recover true clusterings in situations where we have some information about the generating process.

## 1 Introduction

Clustering is a fundamental component of real-world problems in nearly every computational discipline, probably in large part due to the human tendency to use categorization as a tool for understanding data [2]. Also, clustering removes variations due to noise and replication. The value of clustering can indeed be seen by the ubiquity of the $k$-means algorithm and the vast amount of work on clustering that has followed [19].

Clustering is primarily used for two purposes. First, clusters provide compact approximate density representations for multimodal or difficult-to-describe distributions. Second, clustering is used to recover underlying categories in data. In many real-world problems, data points do actually come from a single unobserved class (e. g., an image pixel corresponding to an object), and we would like to group data points based on which unobserved class they come from. This second purpose motivates this work.

In order to properly describe a clustering problem, we often would like to view the data points as having come from more complex distributions than just a mixture of Gaussians in Euclidean space. For example, if we would like to cluster images while maintaining translation invariance, it is unclear how to view each image as a point in some Euclidean space [4]. In this setting, exemplar-based models are appealing, because they do not require any estimation of latent parameters, which may become difficult as spaces and distributions become more complex and high dimensional. Instead, all that is required to cluster data is a computable pairwise similarity measure between all (or a sparse subset of) pairs of points. It is often more natural to describe the clustering problem in this manner. There is an exemplar-based analog to the standard latent-mean algorithm, $k$-means, known as $k$-medians [10].

While exemplar-based models are appealing because continuous latent parameters need not be estimated, learning reduces to a combinatorial optimization problem of identifying exemplars and assigning points to exemplars. However, recent work has revealed efficient algorithms for exemplar-based clustering. Lashkari and Golland [11] give a convex formulation of an exemplar-based model that does not suffer from the initialization problems normally associated with the $k$-means algorithm. Affinity propagation [5] has been shown to find solutions in a matter of minutes that would take $k$-centers days or weeks to find and that outperform Hierarchical Agglomerative Clustering.

One drawback of existing exemplar-based methods, however, is that the implicit prior distribution over clusterings is not explicitly modeled or well-understood. In affinity propagation, for example, different granularities of clusterings are controlled by a hand-tunable parameter, called a *self-similarity* or *preference*, and there is little theoretical justification for setting this parameter.

We introduce a model that admits flexible priors into an exemplar-based clustering framework, allowing us to express a family of flexible and infinite priors over cluster size distributions.

Since our focus is on recovering structure in data, we are interested in maximum a posteriori (MAP) inference algorithms that give a single, hard assignment as output. We develop a max-product belief propagation algorithm for our new model and show experimentally that if we have information about the generating process, without tuning any parameters in our model we are consistently able to recover true clusterings that have an unknown number of clusters and large variations in cluster sizes within individual data sets.

We further show a practical application of our model, where the priors we develop allow control over image segmentations along an axis that (to our knowledge) has been left unexplored in the image segmentation literature.

## 2 Background

### 2.1 Affinity Propagation

Affinity propagation is a clustering algorithm based on max-product belief propagation that is able to cluster data into an *a priori* unknown number of clusters.

The objective function that affinity propagation tries to maximize is:

$$P(C) = \lim_{b \to \infty} \sum_{i=1}^{N} s(i, c_i) - b \sum_{i=1}^{N} f_i(C)$$

where $C = \{c_1, \ldots, c_N\}, c_i \in \{1, \ldots, N\}$ is the index of the exemplar that point $i$ is assigned to, $s$ is a pre-computed pairwise similarity measure between all pairs of points, and $f_i(C)$ is 1 if there is some $j$ such that $c_j = i$ and $c_i \neq i$, and 0 otherwise. Affinity propagation is an optimized max-product belief propagation algorithm over the factor graph shown in Fig. 1.

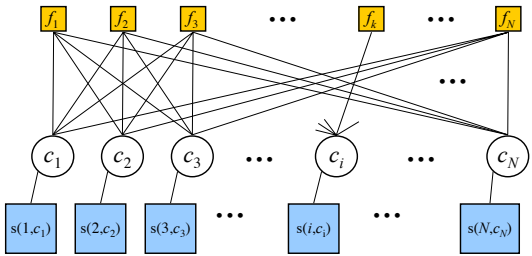

Figure 1: A factor graph representation of affinity propagation.

In affinity propagation, the prior distribution over clusterings, and thus the complexity control setting, is implicit in the user-specified *self similarity* or *preference* for each data point, represented as $s(i, i)$. The self similarity represents each point's tendency to be an exemplar. By making self similarities high, affinity propagation will find a large number of clusters, while making self similarities low will cause affinity propagation to find a small number of clusters. Though in some applications it makes sense to set these self similarities differently for different data points, there is often little reason to favor *a priori* one point to be an exemplar over another. In the latter cases, the self similarities are constrained to take on the same value for all points.

### 2.2 Dirichlet Process Mixture Models

Dirichlet processes provide a well-understood prior over partitions of data, which has been shown to be useful in mixture models used to tackle real-world clustering problems with an *a priori* unknown number of clusters [14]. Dirichlet process mixture models (DPMMs) use a countably infinite number of mixture components in a Bayesian framework to bypass the model selection problem of choosing the number of components [1].

We use the notation $\mathbb{1}[.]$ as the indicator function, which takes a value of 1 if the inside proposition is true and a value of 0 otherwise. For convenience, let $N_k = \sum_{i=1}^{N} \mathbb{1}[c_i = k]$ be the number of points in cluster $k$. Let $K$ be the number of clusters with at least one point. After integrating out mixture weights [8], the probability of a clustering over a set of points $X = \{x_1, \ldots, x_N\}$ being given labels $C = \{c_1, \ldots, c_N\}$ is given as

$$\begin{aligned}
P(C \mid X; G_0, \alpha) &= \prod_{i=1}^{N} P(x_i \mid \theta_{c_i}) \frac{\Gamma(\alpha)}{\Gamma(N + \alpha)} \alpha^K \\
&\quad \times \prod_{k=1}^{K} \Gamma(N_k) \prod_{k=1}^{K} P(\theta_k; G_0)
\end{aligned}$$

where $c_i$ is the cluster assignment for point $i$, $\alpha$ is the concentration parameter, $\theta_j$ are the latent parameters for cluster $j$, and $G_0$ is the base distribution.

In qualitative terms, Dirichlet process mixture models and affinity propagation behave similarly, in that they will continue to find more clusters in data as more data points are observed for a fixed setting of model parameters.

## 3 A Dirichlet Process Exemplar Model

We first develop a Dirichlet process mixture model that uses exemplars instead of latent means. We work in a *collapsed space* (i.e., where mixture weights are integrated out).

Let $X = \{X_e, X_p\}$ where $X_e$ is the set of all points that are exemplars and $X_p$ is the set of all points that are not exemplars. $E = \{e_1, \ldots, e_N\}$ is a set of binary variables, where $e_i = 1$ if point $i$ is an exemplar for its cluster and 0 otherwise. The generative model is then given as follows:

- Draw a partition from a Dirichlet process prior. After

integrating out mixture weights [8], we obtain

$$P(C; \alpha) = \frac{\Gamma(\alpha)}{\Gamma(N+\alpha)} \alpha^K \prod_{k=1}^{K} \Gamma(N_k)$$

- Choose exemplars uniformly at random, but constrain there to be exactly one exemplar per group. The structure over $e$'s is an Markov Random Field with $K$ fully connected cliques involving all points that share the same label:

$$P(E \mid C) = \prod_{k=1}^{K} \frac{1}{Z_k} \textit{one-of-N}(E_k)$$

where $one\text{-}of\text{-}N(E)$ is 1 if exactly one $e \in E$ is 1 and all the rest are zero, and zero otherwise. Let $E_k$ be the set of all $e_i$ such that $c_i = k$. Since there is exactly one legal choice of $e$'s for each choice of exemplar in a group, the partition function $Z_k$ is $N_k$. $P(E \mid C)$ can then be rewritten

$$P(E \mid C) = \prod_{k=1}^{K} \frac{1}{N_k}$$

as long as each non-empty group has exactly one exemplar (i.e., $\forall k \mid N_k > 0, (\sum_{i': c_{i'}=c_k} \mathbb{1}[e_{i'}=1]) = 1$), and zero otherwise.

- Draw parameters for each exemplar from $G_0$:

$$P(X_e; G_0) = \prod_{i=1}^{N} P(x_i; G_0)^{\mathbb{1}[e_i=1]}$$

- Draw the parameters for each remaining point from a distribution parameterized by the exemplar for its group. $P(X_p \mid X_e, C, E)$ is then given as:

$$\prod_{i=1}^{N} \prod_{j=1, j \neq i}^{N} P(x_i \mid x_j)^{\mathbb{1}[c_i=c_j \wedge e_i=0 \wedge e_j=1]}$$

Note that in this model, $x$'s are *not* drawn i.i.d.

The full joint likelihood, $P(C, E, X; G_0, \alpha)$ is

$$P(C; \alpha) P(E \mid C) P(X_e; G_o) P(X_p \mid X_e, C, E)$$

$$= \frac{\Gamma(\alpha)}{\Gamma(N+\alpha)} \alpha^K \prod_{k=1}^{K} \frac{\Gamma(N_k)}{N_k} \prod_{i=1}^{N} P(x_i; G_0)^{\mathbb{1}[e_i=1]}$$

$$\times \prod_{i=1}^{N} \prod_{j=1, j \neq i}^{N} P(x_i \mid x_j)^{\mathbb{1}[c_i=c_j \wedge e_i=0 \wedge e_j=1]}$$

**s.t.** $\forall k \mid N_k > 0, \sum_{i':c_{i'}=k} \mathbb{1}[e_{i'}=1] = 1$

Since the labels $C$ have no meaning beyond saying that points with the same label belong to the same group, we have some freedom in how to choose their values. In particular, we can constrain all groups to take on the label of their exemplar under some fixed but arbitrary ordering of points: $\forall i. c_i = i \iff e_i = 1$. Under this constraint, there is one choice of $C$ for each legal combination of partition and set of exemplars. We can replace all references to $e_i$ with $\mathbb{1}[c_i = i]$. The full likelihood $P(C, X; G_0, \alpha)$ then becomes

$$= \frac{\Gamma(\alpha)}{\Gamma(N+\alpha)} \alpha^{\sum_{i=1}^{N} \mathbb{1}[c_i=i]} \prod_{k:c_k=k} \frac{\Gamma(N_k)}{N_k}$$

$$\times \prod_{i=1}^{N} P(x_i; G_0)^{\mathbb{1}[c_i=i]} P(x_i \mid x_{c_i})^{\mathbb{1}[c_i \neq i]}$$

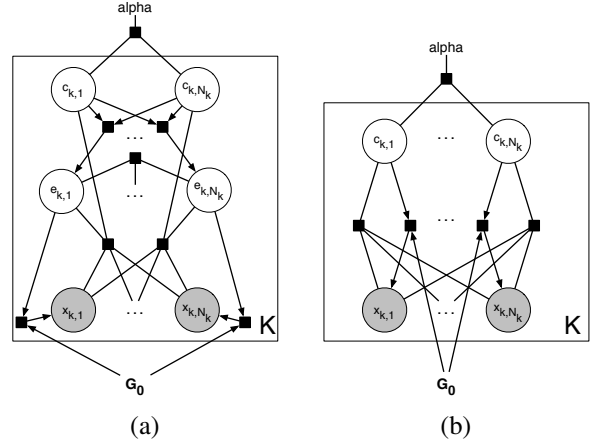

(a)     (b)

Figure 2: (a) A factor graph plate model of the Dirichlet process affinity propagation generative model. (b) After replacing constraints and reparameterizing to remove E.

**s.t.** $\forall k \mid N_k > 0, \quad c_k = k$

Note that under this representation, the labels are not assumed to range from 1 to $K$. Instead, we use not necessarily consecutive labels from 1 to $N$ since given $N$ observed data points, there are at most $N$ clusters in the data. However, we note that we are not explicitly truncating the model by forcing these to correspond to the first $N$ clusters in the stick breaking representation like in [3], and there is no ordering of clusters. In this sense, we are working in the *infinite* Chinese Restaurant Process representation, rather than in a truncated approximation.

### 3.1 Comparison to Affinity Propagation

Affinity propagation uses the same model, but without the Dirichlet process prior over partitions, and rather than drawing exemplar parameters from a given base distribution, self similarities are set as desired by the user.

## 4 Dirichlet Process Affinity Propagation

In order to derive Dirichlet process affinity aropagation (DPAP), a max-product belief propagation algorithm for this model, we make one further change in representation that is useful for deriving extensions to affinity propagation [7]. Rather than representing each $c_i$ as a multinomial variable with $N$ states, we use $N$ binary variables, $\{h_{i1}, \ldots, h_{iN}\}$ and a *one-of-N* constraint specifying that $h_{ij}$ can only be 1 for one choice of $j$. Formally, $c_i = j \Leftrightarrow h_{ij} = 1$. The same algorithm can be derived without making this change of representation, but the derivations are simpler in this form.

By laying out these $h$ variables in a 2-dimensional grid, we can express our model as a factor graph with one factor for each row, one factor for each column, and one factor for each $h$. Fig. 3 shows this factor graph representation that our algorithm operates on.

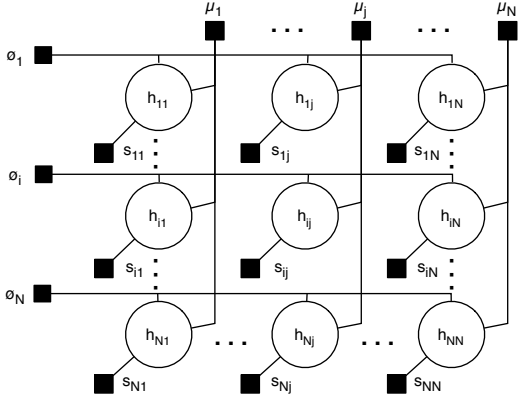

Figure 3: A factor graph representation of the Dirichlet process affinity propagation model as a grid of binary indicator variables.

$$\phi_i(h_{i1}, \ldots, h_{iN}) = \mathbb{1}[\sum_{j=1}^N h_{ij} = 1]$$

$$\mu_j(h_{1j}, \ldots, h_{Nj}) = \begin{cases} 1 & \text{if } N_j = 0 \\ \frac{\Gamma(N_j)}{N_j} \cdot \mathbb{1}[h_{jj} = 1] & \text{otherwise} \end{cases}$$

$$s_{ij}(h_{ij}) = \begin{cases} P(x_i \mid x_j)^{h_{ij}} & \text{if } i \neq j \\ (\alpha \cdot P(x_j; G_0))^{h_{jj}} & \text{if } i = j \end{cases}$$

It can be confirmed that this is equivalent to our earlier formulation.

### 4.1 Max-Product Belief Propagation

Max-product belief propagation is an iterative, local, message passing algorithm that can be used to find the MAP configuration of a discrete probability distribution specified by a factor graph. The algorithm was originally developed for exact inference on tree-structured graphical models, but it has empirically been shown to perform well even on graphs with cycles.

When working in log space, the algorithm is known as max-sum, and the updates for factor graphs involve either a message from a variable to each adjacent factor or a factor to each adjacent variable. The messages from variable to factor add together the messages from all adjacent factors except the factor receiving the message. Formally, if $n(x)$ is the set of all factors that share an edge with $x$, then the message from $x$ to factor $i$, $f_i$, is

$$\tilde{m}_{x \to f_i}(x) = \sum_{f' \in n(x) \setminus f_i} m_{f' \to x}(x)$$

Messages from factors to variables involve a maximization over all variables in the scope of the factor except the variable receiving the message. If $X = neighbors(f)$, then

$$\tilde{m}_{f \to x}(x) = \max_{X \setminus x} \left[ \log f(X) + \sum_{x' \in X \setminus x} m_{x' \to f}(x') \right]$$

We work in log space, and since all variables are binary, we normalize all messages so that $m_{X \to Y}(0) = 0$. This is equivalent to saying that all messages we pass are $m_{X \to Y}(1) = \tilde{m}_{X \to Y}(1) - \tilde{m}_{X \to Y}(0)$.

#### 4.1.1 $\mu$ Factor Messages

The non-trivial calculation that is needed to do max-product inference in this factor graph is to compute the outgoing messages from the $\mu$ factors, $\tilde{m}_{\mu_j \to h_{ij}}(h_{ij})$. We use the notation $h_{-ij} = \{h_{i'j}\}_{i'=1, i' \neq i}^N$ and $h_{i-j} = \{h_{ij'}\}_{j'=1, j' \neq j}^N$ throughout:

$$= \max_{h_{-ij}} \left[ \log \mu_j(h_{:j}) + \sum_{i': i' \neq i}^N m_{h_{i'j} \to \mu_j}(h_{i'j}) \right]$$

$$= \max_{h_{-ij}} \begin{cases} 0 & \text{if } N_j = 0 \\ \log \frac{\Gamma(N_j)}{N_j} + \log \mathbb{1}[h_{jj} = 1] + \\ \sum_{i': i' \neq i}^N m_{h_{i'j} \to \mu_j}(h_{i'j}) & \text{otherwise} \end{cases}$$

If $i = j$:

$$\tilde{m}_{\mu_j \to h_{jj}}(0) = 0$$

$$\tilde{m}_{\mu_j \to h_{jj}}(1) = \max_{h_{-jj}} \sum_{i': i' \neq j}^N m_{h_{i'j} \to \mu_j}(h_{i'j}) +$$

$$\log \frac{\Gamma(1 + \sum_{i': i' \neq j}^N h_{i'j})}{1 + \sum_{i': i' \neq j}^N h_{i'j}}$$

If $i \neq j$:

$$\tilde{m}_{\mu_j \to h_{ij}}(0) = \max(0, \max_{h_{-ij}} \sum_{i': i' \neq i, i' \neq j}^N m_{h_{i'j} \to \mu_j}(h_{i'j}) +$$

$$m_{h_{jj} \to \mu_j}(1) + \log \frac{\Gamma(1 + \sum_{i':i' \neq j, i' \neq i}^{N} h_{i'j})}{1 + \sum_{i':i' \neq j, i' \neq i}^{N} h_{i'j}})$$

$$\tilde{m}_{\mu_j \to h_{ij}}(1) = \max_{h_{-ij}} \sum_{i':i' \neq i, i' \neq j}^{N} m_{h_{i'j} \to \mu_j}(h_{i'j}) +$$

$$m_{h_{jj} \to \mu_j}(1) + \log \frac{\Gamma(2 + \sum_{i':i' \neq j}^{N} h_{i'j})}{2 + \sum_{i':i' \neq j}^{N} h_{i'j}}$$

Temporarily ignoring constants, which are irrelevant in computing the maximal settings of $h$'s, all of the messages require a maximization of the following form:

$$\max_{h_{-ij}} \sum_{i'} m_{h_{i'j} \to \mu_j}(h_{i'j}) + \log \frac{\Gamma(\sum_{i'} h_{i'j})}{\sum_{i'} h_{i'j}}$$

This can be rewritten, using the fact that $h$'s are binary variables, as

$$\max_{h_{-ij}} \sum_{i'} h_{i'j} \cdot m_{h_{i'j} \to \mu_j}(1) +$$
$$(1 - h_{i'j}) \cdot m_{h_{i'j} \to \mu_j}(0)) + \log \frac{\Gamma(\sum_{i'} h_{i'j})}{\sum_{i'} h_{i'j}}$$
$$= \max_{h_{-ij}} \sum_{i'} h_{i'j} \cdot m_{h_{i'j} \to \mu_j}(1) + \log \frac{\Gamma(\sum_{i'} h_{i'j})}{\sum_{i'} h_{i'j}}$$

We can now see that there is a tradeoff between the first term, which specifies how much a point prefers (or prefers not) to be in cluster $j$, and the second term, which favors more points in cluster $j$, regardless of how good of a fit they are. Also notice that as more points are added to cluster $j$, the marginal effect of the second term becomes stronger.

We can effectively remove the $\log \frac{\Gamma(K)}{K}$ term by breaking up the maximization into cases, doing the maximization for each setting of $K = 1, \ldots, N$, and then taking the largest value:

$$\tilde{m}_{\mu_j \to h_{ij}}(h_{ij}; K) + const$$
$$= \log \frac{\Gamma(K)}{K} + \max_{h_{-ij}} \sum_{i'} h_{i'j} m_{h_{i'j} \to \mu_j}(1)$$
$$\textbf{s.t. } \sum_{i'} h_{i'j} = K$$

At this point, it is easy to see that the maximum can be achieved by sorting the $m_{h_{i'j} \to \mu_j}(1)$ values in descending order, and then setting the first $K$ $h_{ij}$'s to be 1 and the remainder to be 0. We do this for each value of $K$, and then take the setting of $K$ that produces the largest value. The sort operation dominates the complexity, so computing a message takes $O(N \log N)$ time, which is an improvement over the $O(2^{N-1})$ time that would be needed to compute the message naively.

By sorting $m_{h_{i'j} \to \mu_j}(1)$ for all values of $i$, rather than for $i' : i' \neq i$, the sort operation can be shared across $N$ instances of essentially the same computation, reducing the complexity to $O(N^2)$ for $N$ of these maximizations that correspond to computing all outgoing messages from a single $\mu$ factor.

Intuitively, this computation is leveraging the fact that the only interaction between the $h$ variables is via the counting term, $\log \frac{\Gamma(K)}{K}$. By conditioning on $K$, the terms break apart and we can maximize greedily. It should be noted that this is similar to the algorithm described in [9], though we are using it in a different way.

#### 4.1.2 $\phi$ Factor Messages

The $\phi$ factor messages specify the $one\text{-}of\text{-}N$ constraint and can be calculated as follows:

$$\tilde{m}_{\phi_i \to h_{ij}}(h_{ij}) = \max_{h_{i-j}} \sum_{j':j' \neq j}^{N} h_{ij'} m_{h_{ij'} \to \phi_i}(h_{ij'})$$
$$\textbf{s.t. } \sum_{j'=1}^{N} h_{ij'} = 1$$
$$\tilde{m}_{\phi_i \to h_{ij}}(1) = 0$$
$$\tilde{m}_{\phi_i \to h_{ij}}(0) = \max_{j':j' \neq j} m_{h_{ij'} \to \phi_i}(1).$$

The message that we actually send is the difference:

$$m_{\phi_i \to h_{ij}}(1) = - \max_{j':j' \neq j} m_{h_{ij'} \to \phi_i}(1).$$

The rest of the messages are computed using standard max-product updates and do not require any marginalization.

#### 4.1.3 Computing Assignments

To compute the belief for $h_{ij}$, we take the standard sum of incoming messages:

$$b_{h_{ij}}(h_{ij}) = s_{ij}(h_{ij}) + m_{\mu_j \to h_{ij}}(h_{ij}) + m_{\phi_i \to h_{ij}}(h_{ij}).$$

Since we enforce that $m(0) = 0$ for all messages, we set $h_{ij} = 1$ if the belief is greater than 0 and $h_{ij} = 0$ otherwise. Upon convergence, we generally do not have problems with more than one $h_{ij} = 1$ per row. However, in the case when it does happen, we set the $h_{ij}$ with the largest belief to be 1 and all others in the row to be zero. As a final step—to refine the final assignment and to eliminate the possibility of finding an illegal solution—we run one round of iterated conditional modes (ICM) (see below), initialized with the settings for $h_{ij}$ that we find with the technique described in this section.

### 4.2 Flexible Priors

At this point it is worth noting that there is nothing specific to the exact form of the Dirichlet process prior that allows us to compute $\mu$ factor messages efficiently. In particular, since the key step of the computation involves pulling the $\log \frac{\Gamma(K)}{K}$ term outside of the maximization, any unnormalized probability that is dependent only on the cluster size may be put in place here. This allows any of the priors described in [18] to be used in place of the Dirichlet process.

## 5 Iterated Conditional Modes

An alternative to max-product inference is to use ICM. For the purposes of this algorithm, we return to the original, expanded formulation of the model, since resampling binary variables with mutual exclusion constraints over rows would be problematic and because a sampler will mix better if it is given more flexibility in the values that labels take on. After proposing a new label for point $i$, we choose the best exemplars for each group, given the new labels. We then take the new label with the largest probability.

We schedule inference in a blocked manner: we iterate over each point, jointly choosing both a new value for the point currently being resampled and a new exemplar for the old and new groups simultaneously. We loop over variables to resample sequentially, until the algorithm converges.

## 6 Experiments

### 6.1 Synthetic Data

We begin by generating synthetic data from the generative process described in section 3. For these experiments, we set $G_0$ to be a spherical Gaussian with unit variance; $P(x_i \mid x_j) = \mathcal{N}(x_i \mid x_j, .5)$; $\alpha = 1$; and we generate 1000 data sets of 100 points each.

We compute $s_{ij}$ for all pairs of points using the distributions given above and run three families of algorithms:

- Affinity propagation (AP(d)): We give standard AP the similarities, $s_{ij}$, as input, along with an additional self-similarity scaling parameter $d$, which is a real number that is added to each $s_{ii}$ before running inference.

- Iterated conditional modes: We initialize the assignment to either one large group (ICM-1) or $N$ separate groups (ICM-N), then run ICM until convergence.

- Dirichlet process affinity propagation (DPAP): The max-product inference algorithm described in section 4.

We scheduled messages in DPAP using a block synchronous schedule, where we alternated between updating mutual exclusion-based messages and cluster-based messages. We determined convergence by checking whether the largest absolute message difference between the current and previous iteration was less than $10^{-5}$. DPAP converged on 94% of the runs, and produced reasonable segmentations even in the few cases when it did not converge. We used message damping of .7 for all cluster-based messages and no damping for mutual exclusion-based messages. Affinity propagation converged on all runs, and we gave it message damping of .8.

We analyze the results of running the algorithms on the synthetic data along several dimensions. First, we look at the distribution of cluster sizes found by each algorithm, aggregated over all the data sets. Fig. 4 (a) shows the distribution of cluster sizes in the true labels for the synthetic data. Fig. 4 (b) shows the distribution of cluster sizes found by affinity propagation when it is given different settings for the diagonal. This corresponds to the range of cluster size distributions that are attainable by adjusting the affinity propagation self-similarity parameter. No matter what settings are chosen for $d$, the algorithm is not able to simultaneously capture the heavy tail of the true labels and the large weight on small clusters of 1-5 points shown in Fig. 4 (a). Fig. 4 (c) shows that adding the infinite prior over cluster sizes allows all of the DPAP model inference algorithms to match the characteristics of the true distribution.

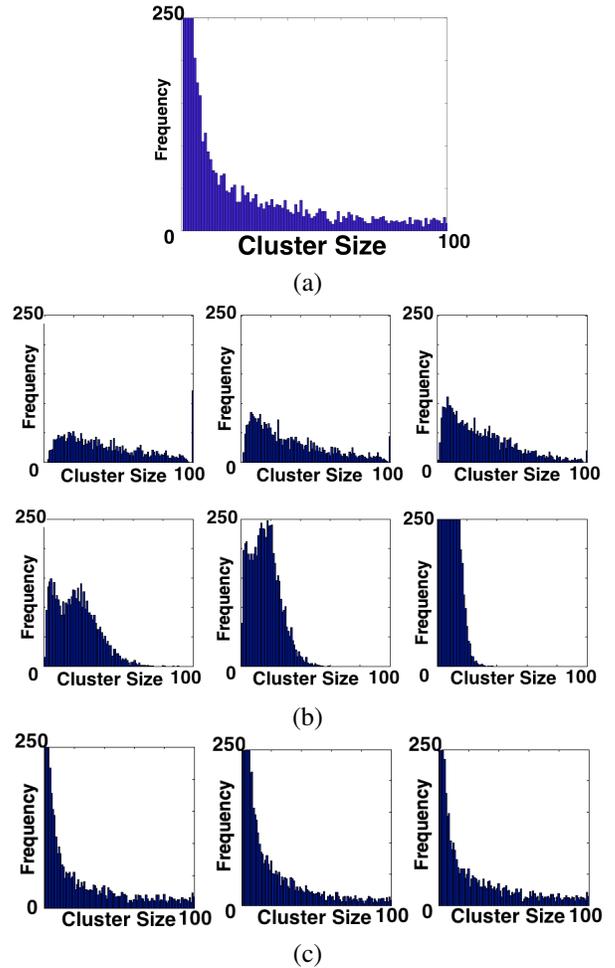

Figure 4: (a) Frequency versus true cluster size for data generated by the exemplar-based generative model, aggregated over 1000 random data sets. (b) Frequency versus size of cluster found by AP($d$) with $d = -100, -50, -35; -20, -10, 0$ (c) Frequency versus size of cluster found by ICM-1, ICM-N, and DPAP, respectively.

Second, we do a comparison with respect to the log likelihoods of the different algorithms that operate on our DPAP model. Figure 5 (a) shows the average delta log like-lihood of the MAP assignment found by each algorithm, as compared to the log likelihood of the true assignment, which is normalized to be the x-axis. Note that all algorithms except for ICM-N consistently find assignments with higher log likelihoods than the true labels do, which is possible due to the random nature of the generating process.

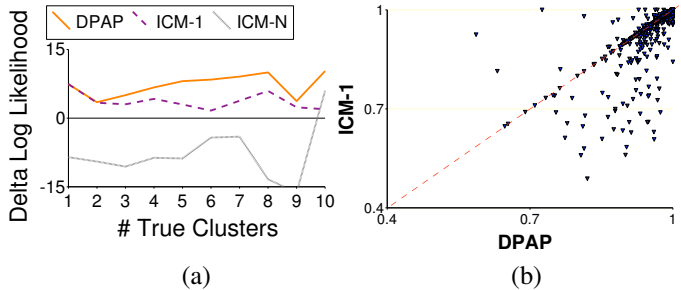

Figure 5: (a) Comparison of log likelihoods of found labels versus number of clusters in the true labels. Scores are normalized by subtracting the likelihood of the true labels. (b) Direct comparison of the top two methods, DPAP and ICM-1, on Rand Index between the true labels and the labels found by each algorithm.

Finally, we directly compare the two best models from (a), DPAP and ICM-1, using the Rand Index [15] of the found labels against the true labels. The Rand Index is a standard metric for measuring the similarity between two clusterings, which is given as the fraction of pairs of points that are correctly assigned as being in the same cluster or correctly assigned as being in different clusters. A score of 1 means that the labels are identical up to a permutation of label names. Figure 5 (b) shows that both do quite well in general, meaning that we are able to recover the underlying labels in many cases, but DPAP consistently outperforms ICM-1, often by a substantial margin.

### 6.2 Real Data: Image Segmentation

Image segmentation is a clustering problem that is made easier if non-Gaussian similarity measures are used. In fact, many approaches to image segmentation do not use latent-mean based models, due in part to the restrictions that they place on the likelihood functions that may be used.

We begin by converting the image into a superpixel representation [16, 6], which has been shown to reduce the size of the problem without losing much information. We find approximately 250 superpixels for each image. We then follow in the spirit of many image segmentation algorithms and use a combination of color and boundary information to compute pairwise similarities between superpixels.

The color component of our similarity measure is based on the difference between mean colors of superpixels in RGB space:

$$s_R(i,j) = -\tau_R ||\bar{R}(i) - \bar{R}(j)||^2$$

To incorporate boundary information, we first use the local boundary detector described in [12] to find a soft edge map, $E$, where each pixel is assigned a probability that it is an edge pixel. For each pair of neighboring superpixels, we look at the pixels on the boundary, $b$, between superpixel $i$ and superpixel $j$, and then take the average edge response over boundary pixels as the edge distance between the superpixel pair.

$$d_E(i,j) = \tau_E \begin{cases} \frac{\sum_{x,y \in b(i,j)} E(x,y)}{|boundary(i,j)|} & \text{if } |b(i,j)| > 0 \\ \infty & \text{otherwise} \end{cases}$$

To convert edge distances to a similarity measure, we find the shortest paths between all pairs of superpixels relative to $d_E$ using Dijkstra's algorithm as is standard. Negating these shortest-path distances gives us the boundary component of our similarity measure, $s_E(i,j)$.

Finally, our full similarity measure simply adds together the two components:

$$s_{ij}(1) = s_R(i,j) + s_E(i,j)$$

It should be noted that we make no claim that our similarity measure between superpixels is optimal. In fact, we make the opposite claim. This was a relatively simple similarity measure that took little hand-tuning to produce. Instead, our aim is to explore whether imposing a prior over cluster sizes is an axis of control that can produce a more powerful segmentation algorithm, a question rarely considered in the image segmentation literature.

Since the focus of this work is on the prior and not image segmentation, we set the values of $\tau_R$ and $\tau_E$ by hand, making sure only that both sources of information contribute roughly equally to the full similarity measure. We leave the values constant for all experiments presented. We give the same similarity matrix to each of the following algorithms, then show results on a range of reasonable parameter settings for each of the respective algorithms:

- Dirichlet process affinity propagation (DPAP): We use DPAP, as suggested by the synthetic experiments. We set the self similarities to be constant at a level that produces an oversegmentation, then we multiply all similarities (self and non-self) by a scaling parameter. This has the effect of varying the relative strength of the prior versus image information. The concentration parameter, $\alpha$, is set to 1 for all experiments.

- Affinity propagation (AP): We give standard AP the similarities, then vary the self similarity to produce a range of clusterings.

- Normalized cuts [17] (Ncut): We give normalized cuts the same similarity matrix as we give the AP and DPAP, then we vary $K$, the number of clusters that we ask it to find.

Figure 6 (a) shows a range of results found by each algorithm on a family scene. The image is quite crowded and has a range of different true cluster sizes. As can be seen by looking at (a6) (where a1-a12 index the color images in raster scan order), the Dirichlet process prior is beneficial, because it allows large clusters like the Christmas tree to remain in a single cluster while simultaneously finely discriminating in the smaller clusters like the faces and chair in the center of the scene. In order for either Ncut or AP to get this level of fineness, they must oversegment the Christmas tree. Compare (a6) to (a8), (a9), (a11), and (a12).

Figure 6 (b) shows segmentations found by each algorithm on a scuba diving scene. The oversegmentation of the background by both AP and Ncut is made more clear in this case, which is likely due to the implicit prior that favors clusters of roughly the same size. However, we can also see that DPAP is too willing to find small, singleton clusters. This comes from the fact that the Dirichlet process prior becomes more discriminating as the clusters become smaller. Since there are few strongly coherent regions other than the background, DPAP is willing to call nearly all of the non-water regions singletons.

Table 1 shows that the DPAP prior provides a quantitative improvement in the clusterings that we are able to find relative to the Rand Index for the two images shown, but we note that of course there are some images where such a prior is appropriate and some where it is not. The images that we have chosen to show are meant to be representative of the range of solutions that each algorithm is able to produce, rather than focusing only on illustrating the best results.

A further point is that all algorithms are too willing to form small, 1-3 superpixel clusters. This suggests that while the Dirichlet process prior is often more appropriate than the implicit priors of AP and Ncut, an ideal image segmentation prior would shift some of the weight that the Dirichlet process prior places on small clusters to medium-sized clusters for these images. Such a prior could be expressed within our model, as mentioned in Section 4.2, but we leave learning or choosing a prior or set of priors specific to image segmentation as future work.

## 7  Discussion

We have developed a family of priors for exemplar-based clustering that are more flexible in their ability to express prior knowledge about cluster size distributions. If there is reason to believe that cluster sizes follow a certain distribution, such as in word counts, various power-law phenomena, or the size of objects in images, then this gives us the ability to impart that information into an exemplar-based clustering setting. The combination of flexible priors with exemplar-based clustering allows us to naturally represent a broad range of clustering problems. With added flexibility comes added ability to discover underlying structure. The tradeoff that we make to add flexibility in the prior information comes at an order $N$ cost in complexity over affinity propagation, due to the fact that we must condition on all values of $K$ inside the $\mu$ factor message computations. However, many of these computations result in the conclusion that no points should be assigned to the given cluster, so there are likely many directions by which to improve efficiency, including calculating bounds and finding intelligent message schedules.

Table 1: Quantitative Results For Real Data

| Family Christmas | | | Scuba Diver | | |
|---|---|---|---|---|---|
| Figure(Alg.) | K | R.I. | Figure(Alg.) | K | R.I. |
| a3(True) | 21 | 1 | b3(True) | 10 | 1 |
| a4(DPAP) | 19 | .746 | b4(DPAP) | 6 | .734 |
| a5(DPAP) | 30 | .884 | **b5(DPAP)** | **15** | **.773** |
| **a6(DPAP)** | **50** | **.937** | b6(DPAP) | 31 | .697 |
| a7(AP) | 7 | .833 | b7(AP) | 5 | .494 |
| a8(AP) | 19 | .915 | b8(AP) | 16 | .457 |
| a9(AP) | 42 | .919 | b9(AP) | 31 | .377 |
| a10(Ncut) | 7 | .862 | b10(Ncut) | 6 | .424 |
| a11(Ncut) | 19 | .914 | b11(Ncut) | 15 | .395 |
| a12(Ncut) | 41 | .917 | b12(Ncut) | 30 | .379 |

If we have no prior information about cluster sizes, it may be reasonable to default to using methods without explicitly defined priors. On the other hand, those methods have implicit priors, which may be equally unjustified. In general, when trying to recover underlying classes of a known type, it is important to consider what assumptions about cluster sizes should be included in the model.

The results presented here show that the combination of flexible priors with exemplar-based clustering gives a large degree of control over the characteristics of the clusterings that our algorithms produce. Our work suggests other research directions along similar lines. One direction that would be interesting is to incorporate a Dirichlet process prior into the $k$-means component of spectral clustering.

We have illustrated our model using Dirichlet process priors on an image segmentation problem, but we emphasize that our model is broadly applicable to many different priors and applications. Further, though there has been some work revealing that max-product belief propagation may be done efficiently with cardinality-based clique potentials [9] and priors placed on node degrees [13], this is the first work that we are aware of that shows how to use Dirichlet process priors in a max-product belief propagation setting. We are currently exploring other problems where using Dirichlet process priors with max-product inference may be useful.

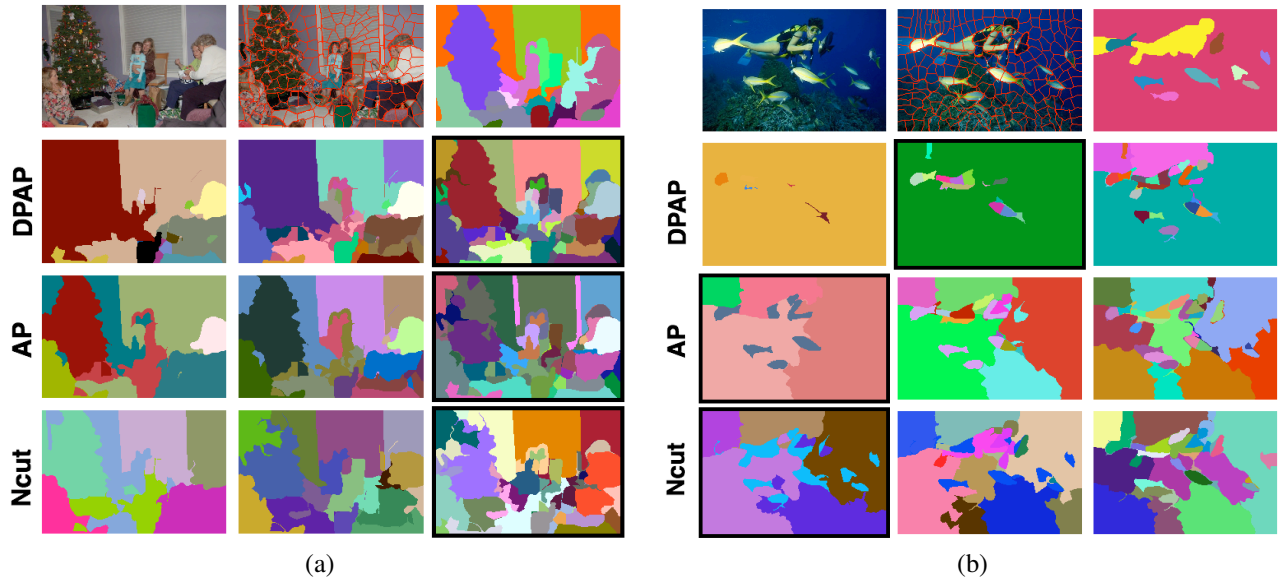

Figure 6: Top row: Original image; Superpixellation; True Labels. Second-Fourth Row: Image segmentations of a crowded family scene and a scuba diving scene using a range of parameter settings for DPAP, AP, and Ncut, respectively. The clustering with the best Rand Index is highlighted in black. Best viewed in color. (a) As AP and Ncut move towards finer segmentations (right column), they oversegment the the Christmas tree. (b) Similarly, as AP and Ncut move towards finer segmentations, they oversegment the water. Since DPAP has an explicit prior over cluster sizes that encourages large clusters to get larger, it chooses to split apart the smaller clusters—the diver and fish—instead of the larger water cluster.